\documentclass[lettersize,journal]{IEEEtran}
\usepackage{amsmath,amsfonts,amssymb}
\usepackage{algorithmic}
\usepackage{algorithm}
\usepackage{array}
\usepackage[caption=false,font=scriptsize,labelfont=sf,textfont=sf]{subfig}
\usepackage{textcomp}
\usepackage{stfloats}
\usepackage{url}
\usepackage{verbatim}
\usepackage{graphicx}
\usepackage{cite}
\usepackage{bm,dsfont}
\usepackage{booktabs} 
\usepackage{array}  
\usepackage{makecell}
\usepackage{subfig}
\usepackage{stackengine}
\usepackage{xcolor}  
\usepackage{colortbl} 
\usepackage{pgf}
\usepackage[table]{xcolor} 
\usepackage[normalem]{ulem}

\usepackage{multirow,booktabs,pifont,siunitx,adjustbox,tikz,xcolor,amssymb}

\usepackage{siunitx} 
\usepackage{multirow} 

\usepackage{xspace}
\makeatletter
\DeclareRobustCommand\onedot{\futurelet\@let@token\@onedot}
\def\@onedot{\ifx\@let@token.\else.\null\fi\xspace}

\def\ie{\emph{i.e}\onedot}

\makeatother

\begin{document}

\title{Aquatic Neuromorphic Optical Flow}

\author{Pei Zhang, Yunkai Liang, and Kaiqiang Wang
        % <-this % stops a space
\thanks{Pei Zhang is with the School of Electrical Engineering, Guangxi University (e-mail: pzhang@gxu.edu.cn).}% <-this % stops a space
\thanks{Yunkai Liang is with the Baise Artificial Intelligence Innovation and Development Center (e-mail: liangyunkai2020@163.com).}% <-this % stops a space
\thanks{Kaiqiang Wang is with the School of Physical Science and Technology, Northwestern Polytechnical University (e-mail: kqwang@nwpu.edu.cn).}% <-this % stops a space
\thanks{Corresponding author: Pei Zhang. This work was supported in part by the Guangxi University Talent Development Funding, and in part by the research funding of Key Laboratory of Photonic Technology for Integrated Sensing and Communication, Ministry of Education, Guangdong University of Technology.}}



\maketitle

\begin{abstract}
Underwater environments impose severe constraints on conventional imaging systems and demand solutions that balance high-quality sensing with strict resource efficiency. While emerging event cameras offer a promising alternative, their potential in aquatic scenarios remains largely unexplored. Through the lens of neuromorphic vision, this work pioneers the investigation of motion fields that serve as key media for agile underwater perception. Built upon spiking neural networks, we introduce a self-supervised framework to estimate per-pixel optical flow from asynchronous event streams, elegantly bypassing the long-standing bottleneck of underwater data scarcity. Extensive evaluations demonstrate that our method achieves competitive visual and quantitative results against leading techniques while operating with superior computational efficiency. By bridging neuromorphic sensing and aquatic intelligence, this work opens new frontiers for lightweight, real-time, and low-cost perception on resource-constrained underwater edge platforms.
\end{abstract}

\begin{IEEEkeywords}
Event camera, underwater, optical flow, spiking neural network, self-supervised learning.
\end{IEEEkeywords}

\section{Introduction}
\begin{figure}[t]
\centering
\setlength{\fboxsep}{0pt}
\setlength{\fboxrule}{0.5pt}

\subfloat[]{%
    \includegraphics[width=0.23\columnwidth]{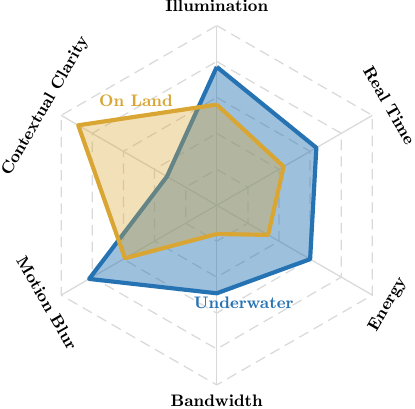}
}
\hfill
\subfloat[]{%
    \includegraphics[width=0.23\columnwidth]{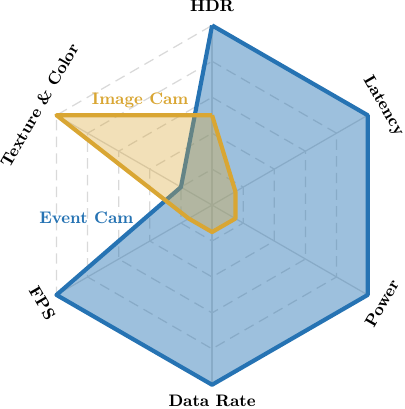}
}
\hfill
\subfloat[]{%
    \begin{tikzpicture}[baseline=(main.south)]
        \node[anchor=south west, inner sep=0pt, outer sep=0pt] (main)
            {\frame{\includegraphics[width=0.23\columnwidth]{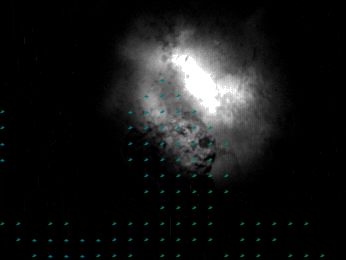}}};
        \node[anchor=south west, inner sep=0pt, outer sep=0pt] at (main.south west) {
            \includegraphics[width=0.05\columnwidth]{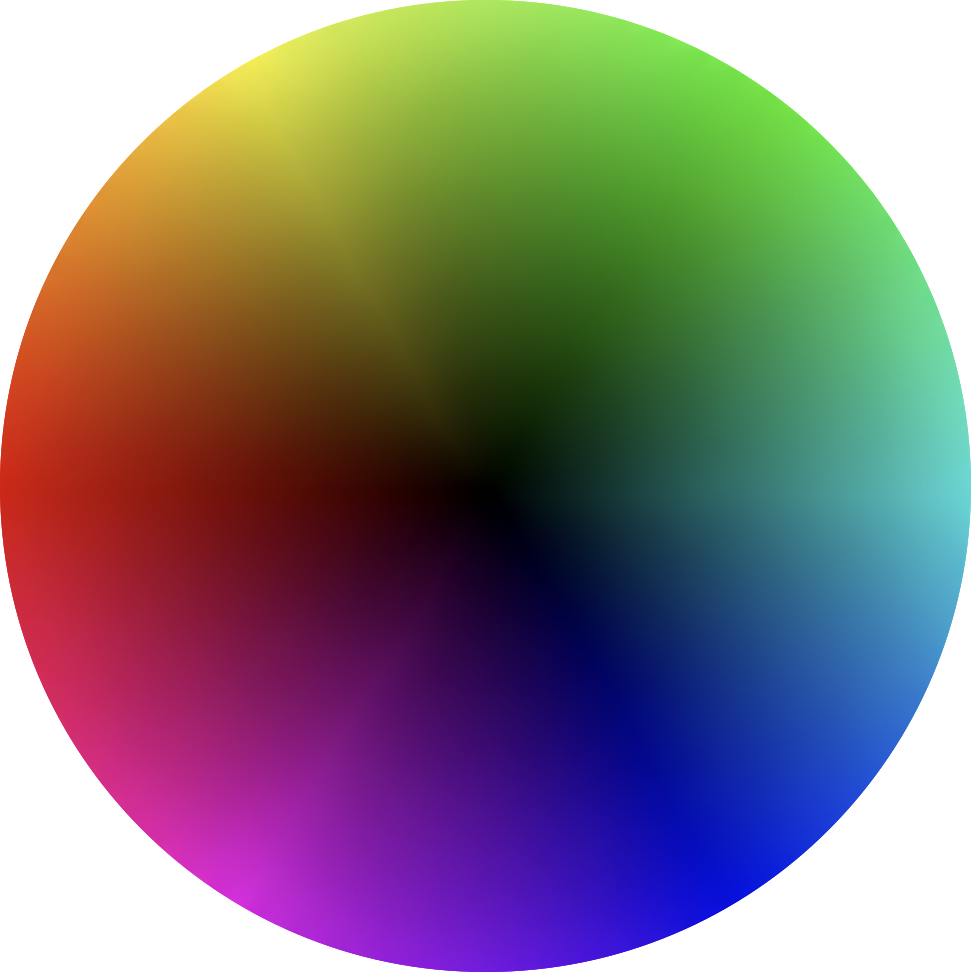}
        };
    \end{tikzpicture}
}
\hfill
\subfloat[]{%
    \begin{tikzpicture}[baseline=(main.south)]
        \node[anchor=south west, inner sep=0pt, outer sep=0pt] (main)
            {\frame{\includegraphics[width=0.23\columnwidth]{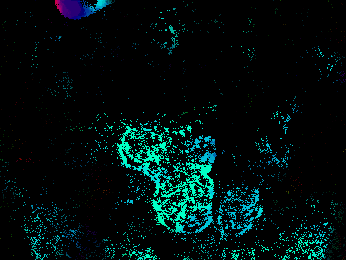}}};
        \node[anchor=north east, inner sep=0pt, outer sep=0pt] at (main.north east) 
            {\includegraphics[width=0.08\columnwidth]{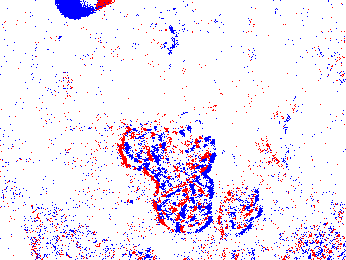}};
    \end{tikzpicture}
}

\caption{(a) Underwater and on-land vision differ in focus. (b) Neuromorphic and frame-based vision differ in features. (c) Frame-based optical flow and a color-coding scheme. (d) Neuromorphic optical flow and source events.}
\label{fig:intro}
\end{figure}

\IEEEPARstart{U}{nderwater} and on-land computer vision diverge in focus and operational demands, with the former contending with unstructured scenes, complex fluid dynamics, low light, and limited onboard computing resources in underwater vehicles~\cite{gonzalez2023survey,zhu2026spectrogen}. A survey in Fig.~\ref{fig:intro}~(a) reveals that underwater efforts are often dedicated to handling challenges such as non-ideal illumination, resource constraints, communication limitations, and degraded imaging quality. Frame-based imaging (\ie, image cameras), while a gold standard on land, becomes a hardware bottleneck in underwater environments. 

Neuromorphic vision via event cameras mimics the human retina and responds to brightness changes with event streams~\cite{schiopu2022lossless,zhang2023neuro,wang2024event,wang2025angle}. As Fig.~\ref{fig:intro}~(b) shows, it enjoys killing features over frame-based vision, including microsecond temporal resolution along with low data rates, $120$~\si{\decibel} high dynamic range (HDR) to overcome backscatter and light attenuation, and minimal power consumption~\cite{zhang2025tcsvt}. These make it a promising solution for resource-constrained mobile platforms. Despite widespread terrestrial success~\cite{gehrig2024low,zhu2025ultrafast,chen2024event,chen2026self,shiba2023fast,wu2026dark}, its underwater use remains emerging. AquaticVision~\cite{peng2025aquaticvision} and OceanLab~\cite{oceanlab} pioneered event-based SLAM benchmarks, yet Table~\ref{tab:cf_dataset} reveals a persistent scarcity of diverse, large-scale datasets with reliable ground truth. Beyond navigation, event cameras enhance AUV docking via fast marker localization~\cite{zhang2022event}, serve as scientific tools to observe millisecond-scale plankton behaviors~\cite{takatsuka2024millisecond} and transparently camouflaged organisms~\cite{luo2023transcodnet}, and boost underwater image restoration~\cite{bi2022ao, bi2024fms}. These works fully cover the current state of research on underwater neuromorphic vision.

In dynamic fluid environments, optical flow is a basic but vital cue for downstream tasks like navigation and SLAM, detection and tracking, imaging reconstruction, and flow-aware control. However, it remains unexplored for the lack of large-scale, annotated event data. Then, event-based self-supervised learning becomes a dominant paradigm practical for real-world deployment. Despite success on land~\cite{zhu2018ev,hagenaars2021nips,paredes2023iccv}, its viability for underwater use remains unverified. Another challenge lies in the tension between limited computing resources on underwater platforms and the resource-intensive nature of standard artificial neural networks (ANNs). Addressing both algorithm and hardware bottlenecks, spiking neural networks (SNNs) offer a convincing solution for their computational efficiency and native capacity to handle spike-like events~\cite{maass1997nn,kosta2023adaptive}, making them well suited for underwater tasks~\cite{li2025yolo,sudevan2025arxiv}.

To our knowledge, this work pioneers the study of underwater motion fields through the lens of neuromorphic vision. We present a self-supervised SNN framework to estimate per-pixel optical flow from event streams. As compared in Fig.~\ref{fig:intro}~(c) and (d), our neuromorphic solution delivers a more impressive result for aquatic scenes by seeing through the dark and blind spots between frames. It earns competitive performance against leading ANN baselines while operating with superior computational efficiency, and also benefits downstream tasks like tracking camouflaged marine life. This work opens new avenues for lightweight, real-time, and low-cost perception on resource-constrained underwater edge systems.
\begin{table*}[t]
\caption{Existing Underwater Neuromorphic Datasets}
\label{tab:cf_dataset}
\centering
\renewcommand{\arraystretch}{0.95} 
\setlength{\tabcolsep}{0pt} 
\begin{tabular*}{\textwidth}{@{\extracolsep{\fill}}lrrrr}
\toprule
 & \textbf{AquaticVision (2025)}~\cite{peng2025aquaticvision}  & \textbf{Aqua-Eye (2024)}~\cite{luo2023transcodnet} & \textbf{OceanLab (2023)}~\cite{oceanlab} & \textbf{DAVIS-NUIUIED (2022)}~\cite{bi2022ao} \\
\midrule
\textbf{Task}          & SLAM     &  object detection   & SLAM        & image enhancement \\ 
\textbf{Sensor}    & $2\times$DAVIS346 & DAVIS346 &  \makecell[r]{DAVIS346\\HD frame camera}& DAVIS346\\ 
\textbf{Pixel}       & $346\times260$ & $346\times260$ & \makecell[r]{$346\times260$\\$1920\times1080$}&$346\times260$ \\ 
\textbf{Annotation}    &  ---      & bounding box      & ---   & ---\\ 
\textbf{Environment}      & lab tank       & natural scene  & natural scene  & lab tank  \\ 
\textbf{Dataset Scale} &  limited & moderate & limited & limited \\ 
% \textbf{Scene Complexity}  & dominant subject & weak dynamics & sparse target &rich texture\\ 
\bottomrule
\end{tabular*}
\end{table*}

\section{Methodology}
\subsection{Self-Supervised Learning for Optical Flow}
\begin{figure*}[t]
    \centering
    \hfill
    \subfloat[]{%
        \includegraphics[width=0.28\textwidth]{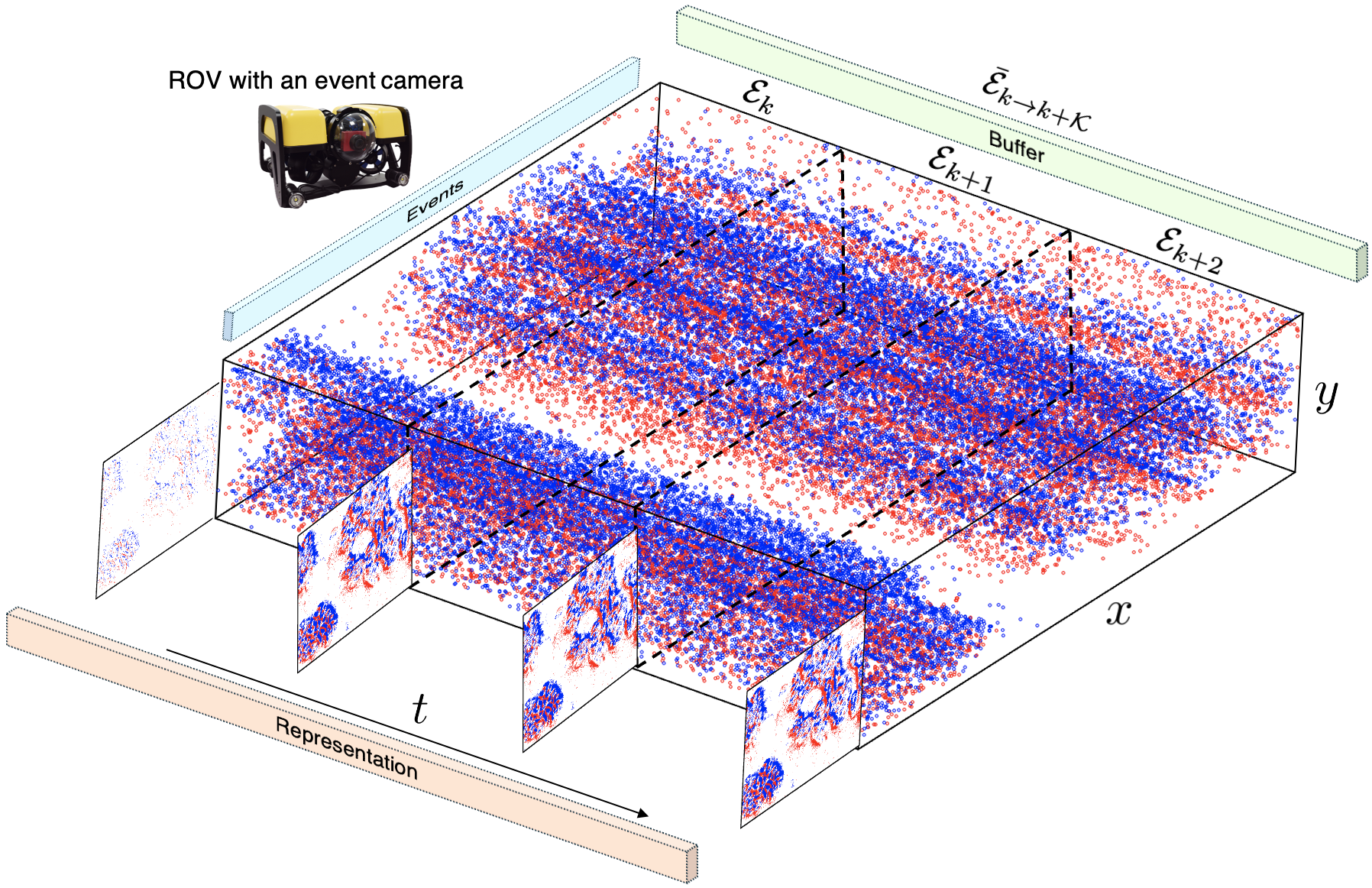}%
    }%
    \hfill
    \subfloat[]{%
        \includegraphics[width=0.36\textwidth]{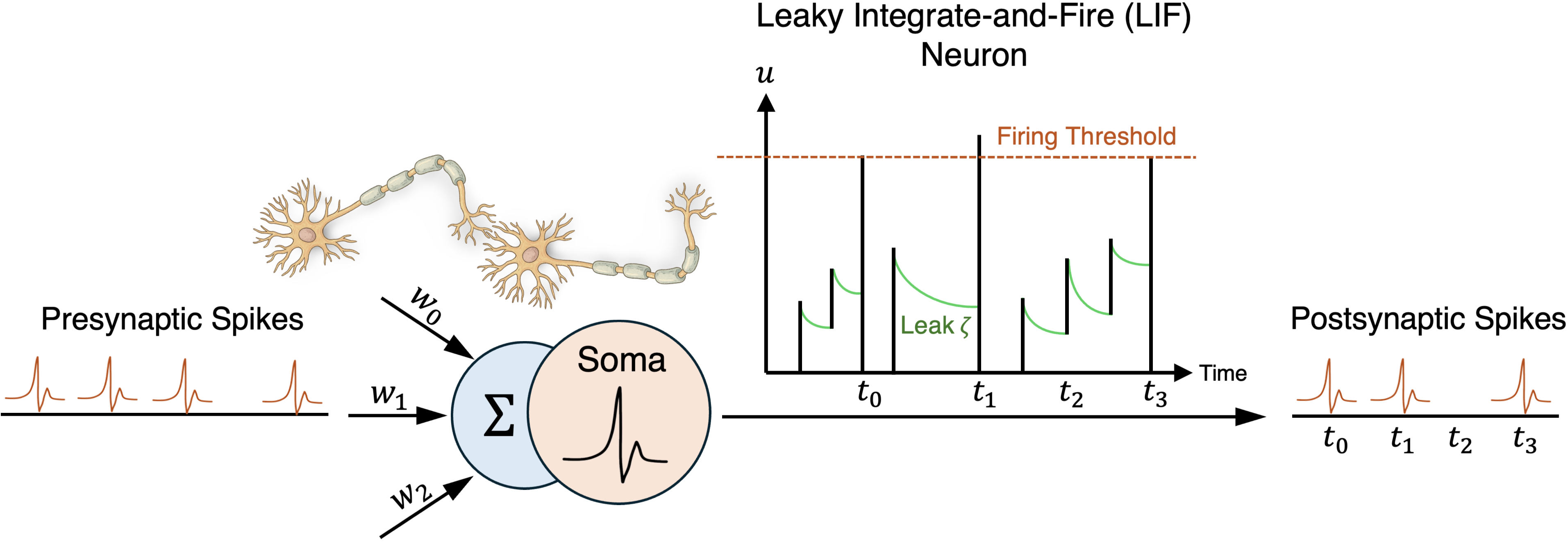}%
    }%
    \hfill
    \subfloat[]{%
        \includegraphics[width=0.32\textwidth]{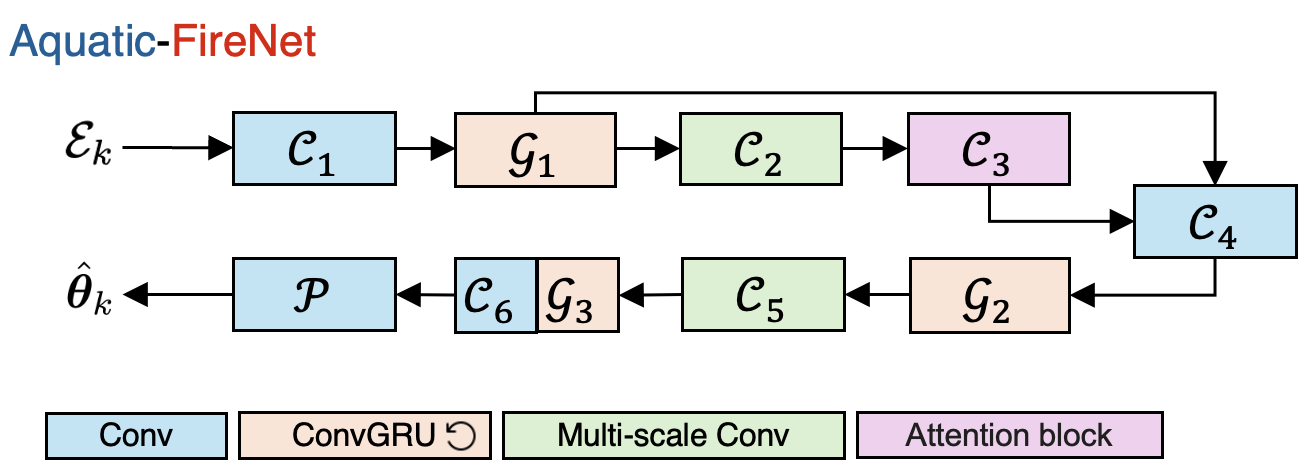}%
    }%
    \hfill
    \caption{(a) An event stream of a coral scene and its corresponding event-count representations. (b) Dynamics of an LIF neuron in which the firing threshold and leak are learnable. (c) Overview of the proposed Aq-FireNet.}
    \label{fig:demo}
\end{figure*}

Underwater perception is degraded by light absorption and scattering, which attenuates visibility, reduces scene contrast, and amplifies sensor noise~\cite{mayerhofer2020}. The light intensity at a spatial position 
$\mathbf{x} = (x, y)^\intercal$ at a depth $z$, is denoted as $I(\mathbf{x}, z) = I_0(\mathbf{x})\mathrm{e}^{-\alpha z}$, with the initial intensity $I_0$ and attenuation factor $\alpha$ anchored on light and water properties. Event cameras detect per-pixel brightness changes and encode an event $e = (\mathbf{x}, t, p)$ triggered at $\mathbf{x}$ at a timestamp $t$ when the variation $\Delta L(\mathbf{x}, z, t)$ over the time elapsed $\Delta t$ reaches a contrast threshold $c$
\begin{equation}
    \Delta L(\mathbf{x}, z, t) = \log \left(\frac{I(\mathbf{x}, z, t)}{I(\mathbf{x}, z, t-\Delta t)}\right) = pc,
\end{equation}
with the polarity $p\in\{+1,-1\}$ for the change direction. Then, an event stream $\mathcal{E}$, comprising all online events from underwater scenes, is denoted as $\mathcal{E} = \{e_i\}_{i=0:\infty} = \big\{(\mathbf{x}_i, t_i, p_i)\big\}_{i=0:\infty}$, with each being indexed by $i$. Beyond target motion, environmental dynamics such as turbulence and suspended particulates, often trigger background noise. We harness count encoding, a computationally-efficient representation meshing with resource-constrained underwater settings. Here, events are accumulated per-pixel and per-polarity over a time interval to shape event-count images for continuous optical flow estimation~\cite{hagenaars2021nips}. As Fig.~\ref{fig:demo}~(a) shows, each image $k$ is populated with the consecutive, non-overlapping partitions of a stream $\mathcal{E}_k = \{e_i\}_{i=0}^{K-1}$ (\ie, each has $K$ events), ensuring a compact yet precise depiction of dynamic scenes.

We exploit contrast maximization (CM) to estimate per-pixel flow fields from event streams in a self-supervised fashion~\cite{gallego2018cvpr,gallego2019cvpr}. While conventional models assume linear warping in rigid scenes, underwater environments introduce optical distortions, driven by local refractive index variations, to cause magnification and shrinkage of targets. To describe such fluid scenes, we augment the linear model with isotropic scaling and represent motion as a synergy of translation and scaling. Assuming the candidate motion vector $\bm{\theta}(\mathbf{x}) = \big(\mathbf{v}(\mathbf{x}), \phi(\mathbf{x})\big)^\intercal$, where $\mathbf{v} = (v_x, v_y)^\intercal$ is the velocity, is constant within a small spatiotemporal partition, each event is propagated to a reference time $t'$ and a warped position $\mathbf{x}'$
\begin{equation}
\mathbf{x}'_i = \underbrace{\mathbf{x}_i + \mathbf{v}(\mathbf{x}_i)(t' - t_i)}_{\text{Linear Warping}} + \underbrace{\phi(\mathbf{x}_i) (\mathbf{x}_i - \mathbf{x}^c)(t' - t_i)}_{\text{Isotropic Scaling}}.
\end{equation}
The factor $\phi$ is an approximation of the divergence of the local distortion field, and captures the expansion ($\phi>0$) or contraction rate ($\phi<0$). This micro-refraction is assumed isotropic for stable, parameter-efficient optimization. Due to the unknown optical center of the distortion, we set $\mathbf{x}^c$ to the spatial centroid of the given partition. This aligns micro-refraction effects with regions of highest event activity, enabling efficient estimation without learnable parameters.

The aggregation result is termed an image of warped events (IWE). Then, the IWE of the per-pixel average timestamp for each polarity $p'$ at $t'$, denoted as $T_{p'}(\mathbf{x}, \bm{\theta} \mid t')$, is shaped by a linear interpolation kernel $\sigma(x)=\text{max}\big(0, 1- |x|\big)$
\begin{equation}\label{eq:rep}
    T_{p'}(\mathbf{x}, \bm{\theta} \mid t')=\frac{\sum_i \mathds{1}_{p_i=p'} \sigma(x-x'_i)\sigma(y-y'_i)t_i}{\sum_i \mathds{1}_{p_i=p'} \sigma(x-x'_i)\sigma(y-y'_i)+\epsilon},
\end{equation}
with the offset $\epsilon$ for numerical stability. With accurate flow, events from the same moving edge align tightly in space-time after motion compensation, leading to a sharper event distribution and a lower $T_{p'}$. Then, the CM loss $\mathcal{L}_\text{CM}$ at $t'$ is the scaled sum of the squared temporal images (\ie, edge energy) for both polarities across all pixels
\begin{equation}
    \mathcal{L}_\text{CM}(t') = \frac{\sum_\mathbf{x} f(T_+, t') + f(T_-, t')}{\sum_\mathbf{x}\mathds{1}_{\delta(\mathbf{x}')>0}+\epsilon},  
\end{equation}
where $\delta(\mathbf{x}')$ is a per-pixel event count of the IWE for ensuring loss convexity, and the function $f(T,t') = \|T(\mathbf{x},\bm{\theta} \mid t')\|^{2} + \lambda_0 \| T(\mathbf{x},\bm{\theta} \mid t')\|_{0}$. Incorporating the $L_0$ term for noisy event-based reconstruction has been verified~\cite{zhang2024tip}. In underwater scenes where optical degradation induces diffuse event patterns and ambiguity, it complements temporal alignment to penalize minor deviations in the IWE, encouraging sharper motion boundaries while reducing blur and noise. Given its non-convex and non-differentiable nature, we use a differentiable approximation for optimization $\|\mathbf{x}\|_0 \approx \sum_i \tanh^2(|x_i|/\epsilon)$. To mitigate temporal scaling issues during backpropagation~\cite{zhu2019cvpr}, the CM loss is computed bidirectionally to ensure equal contributions from all events $\mathcal{L}_\text{CM} = \mathcal{L}_\text{CM}(t_0) + \mathcal{L}_\text{CM}(t_{K-1})$. At the flow level, a smoothness regularizer $\mathcal{L}_\text{SM}$ penalizes flow variations to enforce spatial collinearity
\begin{equation}
\mathcal{L}_\text{SM}(\bm{\theta}) = \sum_{\mathbf{x}} \sum_{\mathbf{n} \in \mathcal{N}(\mathbf{x})} \left\| \rho \big( \bm{\theta}(\mathbf{x}) - \bm{\theta}(\mathbf{n}) \big) \right\|_1,
\end{equation}
with the Charbonnier loss $\rho(x) = (x^2 + \epsilon^2)^\beta$ for outlier rejection and the set of spatial neighbors $\mathcal{N}(\mathbf{x})$. We set $\beta=0.45$ and $\epsilon=0.001$.
Finally, the overall loss $\mathcal{L}$ to supervise network training is $\mathcal{L} = \mathcal{L}_\text{CM} + \lambda\mathcal{L}_\text{SM}$. As shown in Fig.~\ref{fig:demo}~(a), a buffer $\bar{\mathcal{E}}_{k\rightarrow {k+\mathcal{K}}} \doteq \big\{(\mathcal{E}_i, \hat{\bm{\theta}}_i)\big\}_{i=k}^{k+\mathcal{K}}$ accumulates $\mathcal{K}$ forward-pass event-flow tuples to ensure adequate motion context for optimization. Parameters are then updated via backpropagation through time, followed by network state detachment and buffer resets.

\begin{table}[!t]
    \centering
    \footnotesize
    \setlength{\tabcolsep}{2.3pt}
    \renewcommand{\arraystretch}{1.1}

    \caption{Quantitative Evaluation by Supervised Metrics}
    \label{table/ex_quan}

    \sisetup{detect-weight, mode=text}
    \definecolor{AEE_Best}{HTML}{51A2FF}   
    \definecolor{AEE_Good}{HTML}{8EC5FF}   
    \definecolor{AEE_Avg}{HTML}{BEDBFF}    
    \definecolor{AEE_Poor}{HTML}{DBEAFE}   

    \definecolor{PXL_Best}{HTML}{FDC745}   
    \definecolor{PXL_Good}{HTML}{FFDF20}   
    \definecolor{PXL_Avg}{HTML}{FFF085}   
    \definecolor{PXL_Poor}{HTML}{FEF9C2}   

    \begin{tabular}{@{} l
                     S[table-format=1.2] S[table-format=2.2]
                     S[table-format=1.2] S[table-format=1.2]
                     S[table-format=1.2] S[table-format=1.2]
                     S[table-format=1.2] S[table-format=1.2] @{}}
    \toprule
    & \multicolumn{2}{c}{\texttt{A}~\cite{peng2025aquaticvision}}
    & \multicolumn{2}{c}{\texttt{B}~\cite{luo2023transcodnet}}
    & \multicolumn{2}{c}{\texttt{C}~\cite{luo2023transcodnet}}
    & \multicolumn{2}{c}{\texttt{D}~\cite{bi2022ao}} \\
    \cmidrule(lr){2-3} \cmidrule(lr){4-5} \cmidrule(lr){6-7} \cmidrule(lr){8-9}
    & {AEE} & {$\%_\text{3px}$}
    & {AEE} & {$\%_\text{3px}$}
    & {AEE} & {$\%_\text{3px}$}
    & {AEE} & {$\%_\text{3px}$} \\
    \midrule
    FireNet         & \cellcolor{AEE_Poor}1.57 & \cellcolor{PXL_Poor}13.39 & \cellcolor{AEE_Good}1.22 & \cellcolor{PXL_Poor}4.72 & \cellcolor{AEE_Avg}0.43 & \cellcolor{PXL_Good}0.21 & \cellcolor{AEE_Avg}0.70 & \cellcolor{PXL_Best}1.22 \\
    EV-FlowNet      & \cellcolor{AEE_Good}1.41 & \cellcolor{PXL_Best}9.28  & \cellcolor{AEE_Best}1.05 & \cellcolor{PXL_Best}3.39 & \cellcolor{AEE_Best}0.41 & \cellcolor{PXL_Good}0.19 & \cellcolor{AEE_Best}0.62 & \cellcolor{PXL_Best}0.92 \\
    \addlinespace
    LIF-FireNet     & \cellcolor{AEE_Poor}1.63 & \cellcolor{PXL_Poor}15.02 & \cellcolor{AEE_Poor}1.38 & \cellcolor{PXL_Good}3.45 & \cellcolor{AEE_Poor}0.50 & \cellcolor{PXL_Poor}0.35 & \cellcolor{AEE_Poor}0.71 & \cellcolor{PXL_Poor}2.03 \\
    XLIF-FireNet    & \cellcolor{AEE_Avg}1.48 & \cellcolor{PXL_Good}11.13 & \cellcolor{AEE_Avg}1.29 & \cellcolor{PXL_Avg}4.31 & \cellcolor{AEE_Poor}0.46 & \cellcolor{PXL_Poor}0.33 & \cellcolor{AEE_Good}0.68 & \cellcolor{PXL_Poor}1.95 \\
    LIF-EV-FlowNet  & \cellcolor{AEE_Avg}1.52 & \cellcolor{PXL_Avg}11.79 & \cellcolor{AEE_Avg}1.31 & \cellcolor{PXL_Good}3.52 & \cellcolor{AEE_Poor}0.46 & \cellcolor{PXL_Poor}0.56 & \cellcolor{AEE_Good}0.63 & \cellcolor{PXL_Good}1.36 \\
    XLIF-EV-FlowNet & \cellcolor{AEE_Best}1.40 & \cellcolor{PXL_Good}10.74 & \cellcolor{AEE_Good}1.18 & \cellcolor{PXL_Best}3.12 & \cellcolor{AEE_Avg}0.43 & \cellcolor{PXL_Good}0.20 & \cellcolor{AEE_Best}0.60 & \cellcolor{PXL_Good}1.30 \\ \addlinespace
    Aq-FireNet      & \cellcolor{AEE_Best}1.40 & \cellcolor{PXL_Best}9.92  & \cellcolor{AEE_Good}1.20 & \cellcolor{PXL_Best}3.33 & \cellcolor{AEE_Best}0.41 & \cellcolor{PXL_Best}0.18 & \cellcolor{AEE_Good}0.68 & \cellcolor{PXL_Avg}1.49 \\
    Aq-FireNet ($\phi=0$) & \cellcolor{AEE_Good}1.43 & \cellcolor{PXL_Avg}11.21 & \cellcolor{AEE_Avg}1.26 & \cellcolor{PXL_Good}3.74 & \cellcolor{AEE_Good}0.42 & \cellcolor{PXL_Avg}0.23 & \cellcolor{AEE_Avg}0.70 & \cellcolor{PXL_Avg}1.61\\
    Aq-FireNet ($\lambda_0=0$) & \cellcolor{AEE_Avg}1.48 & \cellcolor{PXL_Avg}12.37 & \cellcolor{AEE_Poor}1.32 & \cellcolor{PXL_Avg}3.97 & \cellcolor{AEE_Avg}0.44 & \cellcolor{PXL_Avg}0.27 & \cellcolor{AEE_Avg}0.70 & \cellcolor{PXL_Poor}1.80\\
    \bottomrule
\end{tabular}

    \begin{minipage}{\columnwidth}
        \vspace{2pt}
        \footnotesize
         \texttt{A. scan\_with\_board \hfill C. moon\_jellyfish2 \\ B. sum10 \hfill D. scene\_3}\\
         Color encodes the scores, the darker the better.
    \end{minipage}

\end{table}

\subsection{Neuron Model and Network Architecture}
As Fig.~\ref{fig:demo}~(b) illustrates, the LIF neuron integrates incoming synaptic input over time with a leakage term modeling membrane decay~\cite{lapicque1997brain}. The membrane potential $u_i[t]$ of a postsynaptic neuron $i$ at a timestep $t$ follows
\begin{equation}
    u_i[t] = \zeta u_i[t-1] + (1-\zeta )\Big(\sum_j \tilde{w}_{ij} s_j[t] + \sum_o \check{w}_{io} s_o[t-1]\Big),
\end{equation}
where $\zeta$ is the membrane decay, and $\sum_j \tilde{w}_{ij} s_j[t]$, with the feedforward weight $\tilde{w}$, represents the total synaptic input at the current timestep, computed as the weighted sum of incoming spikes (\ie, events) $s_j[t] \in \{0,1\}$ from all presynaptic neurons $j$. Similar explanation for the recurrent weight $\check{w}$ and the index $o$. When the potential exceeds a firing threshold, the neuron emits a spike and then resets. Repeated over time, this process enables event-driven computation across SNN layers.

As Fig.~\ref{fig:demo}~(c) depicts, our prototype Aquatic-FireNet (Aq-FireNet) inherits a topology from FireNet~\cite{scheerlinck2020wacv} for its fast, lightweight, and computationally-efficient features, to match real-time camera responses and resource-constrained underwater platforms. To handle turbidity-induced scale variations and visual distortions, we leverage a dual-pathway, multi-scale convolutional architecture augmented by cross-pathway skip connections. A lightweight spiking attention module is integrated to amplify salient motion while suppressing background noise. Spatiotemporal modeling is achieved through a hybrid recurrent strategy that couples the implicit short-term memory of LIF neurons with explicit long-term ConvGRUs~\cite{ballas2015iclr}.

\begin{table}[!t]
    \centering
    \footnotesize
    \setlength{\tabcolsep}{2.3pt}
    \renewcommand{\arraystretch}{1.1}

    \caption{Quantitative Evaluation by Unsupervised Metrics}
    \label{table/ex_quan2}

    \sisetup{detect-weight, mode=text}

    \definecolor{FWL_Best}{HTML}{51A2FF} 
    \definecolor{FWL_Good}{HTML}{8EC5FF} 
    \definecolor{FWL_Avg}{HTML}{BEDBFF}  
    \definecolor{FWL_Poor}{HTML}{DBEAFE} 

    \definecolor{RSAT_Best}{HTML}{FDC745} 
    \definecolor{RSAT_Good}{HTML}{FFDF20} 
    \definecolor{RSAT_Avg}{HTML}{FFF085}  
    \definecolor{RSAT_Poor}{HTML}{FEF9C2} 

    \begin{tabular}{@{} l *{8}{S[table-format=1.2]} @{}}
    \toprule
    & \multicolumn{2}{c}{\texttt{A}~\cite{peng2025aquaticvision}}
    & \multicolumn{2}{c}{\texttt{B}~\cite{luo2023transcodnet}}
    & \multicolumn{2}{c}{\texttt{C}~\cite{oceanlab}}
    & \multicolumn{2}{c}{\texttt{D}~\cite{bi2022ao}} \\
    \cmidrule(lr){2-3} \cmidrule(lr){4-5} \cmidrule(lr){6-7} \cmidrule(lr){8-9}
    & {FWL} & {RSAT}
    & {FWL} & {RSAT}
    & {FWL} & {RSAT}
    & {FWL} & {RSAT} \\
    \midrule
    FireNet         & \cellcolor{FWL_Good}1.13 & \cellcolor{RSAT_Best}0.91 & \cellcolor{FWL_Best}1.49 & \cellcolor{RSAT_Good}0.90 & \cellcolor{FWL_Good}1.36 & \cellcolor{RSAT_Avg}1.01 & \cellcolor{FWL_Best}1.82 & \cellcolor{RSAT_Good}0.83\\
    EV-FlowNet      & \cellcolor{FWL_Best}1.23 & \cellcolor{RSAT_Good}0.93 & \cellcolor{FWL_Good}1.46 & \cellcolor{RSAT_Best}0.83 & \cellcolor{FWL_Avg}1.33 & \cellcolor{RSAT_Good}0.97 & \cellcolor{FWL_Avg}1.74 & \cellcolor{RSAT_Best}0.81\\
    \addlinespace
    LIF-FireNet     & \cellcolor{FWL_Poor}1.01 & \cellcolor{RSAT_Poor}1.01 & \cellcolor{FWL_Avg}1.42 & \cellcolor{RSAT_Poor}1.01 & \cellcolor{FWL_Avg}1.32 & \cellcolor{RSAT_Poor}1.04 & \cellcolor{FWL_Poor}1.61 & \cellcolor{RSAT_Poor}0.87\\
    XLIF-FireNet    & \cellcolor{FWL_Poor}1.06 & \cellcolor{RSAT_Good}0.95 & \cellcolor{FWL_Good}1.44 & \cellcolor{RSAT_Avg}0.94 & \cellcolor{FWL_Good}1.35 & \cellcolor{RSAT_Poor}1.03 & \cellcolor{FWL_Poor}1.70 & \cellcolor{RSAT_Avg}0.85\\
    LIF-EV-FlowNet  & \cellcolor{FWL_Avg}1.08 & \cellcolor{RSAT_Avg}0.97 & \cellcolor{FWL_Avg}1.40 & \cellcolor{RSAT_Good}0.91 & \cellcolor{FWL_Poor}1.29 & \cellcolor{RSAT_Avg}0.98 & \cellcolor{FWL_Poor}1.67 & \cellcolor{RSAT_Poor}0.87\\
    XLIF-EV-FlowNet & \cellcolor{FWL_Good}1.13 & \cellcolor{RSAT_Good}0.92 & \cellcolor{FWL_Poor}1.38 & \cellcolor{RSAT_Good}0.91 & \cellcolor{FWL_Poor}1.29 & \cellcolor{RSAT_Good}0.97 & \cellcolor{FWL_Avg}1.73 & \cellcolor{RSAT_Avg}0.84\\ \addlinespace
    Aq-FireNet      & \cellcolor{FWL_Best}1.18 & \cellcolor{RSAT_Good}0.95 & \cellcolor{FWL_Best}1.48 & \cellcolor{RSAT_Avg}0.92 & \cellcolor{FWL_Good}1.36 & \cellcolor{RSAT_Avg}0.98 & \cellcolor{FWL_Good}1.78 & \cellcolor{RSAT_Good}0.83\\
    Aq-FireNet ($\phi=0$)      & \cellcolor{FWL_Good}1.14 & \cellcolor{RSAT_Avg}0.99 & \cellcolor{FWL_Avg}1.43 & \cellcolor{RSAT_Avg}0.92 & \cellcolor{FWL_Good}1.35 & \cellcolor{RSAT_Avg}1.01 & \cellcolor{FWL_Good}1.75 & \cellcolor{RSAT_Avg}0.84\\
    Aq-FireNet ($\lambda_0=0$)      & \cellcolor{FWL_Avg}1.10 & \cellcolor{RSAT_Poor}1.00 & \cellcolor{FWL_Avg}1.42 & \cellcolor{RSAT_Avg}0.94 & \cellcolor{FWL_Good}1.33 & \cellcolor{RSAT_Poor}1.02 & \cellcolor{FWL_Avg}1.71 & \cellcolor{RSAT_Poor}0.86\\
    \bottomrule
\end{tabular}

    \begin{minipage}{\columnwidth}
        \vspace{2pt}
        \footnotesize
         \texttt{A. hdr \hfill C. recording6 \\ B. flame\_jellyfish19  \hfill D. scene\_9}\\
        Color encodes the scores, the darker the better.
    \end{minipage}

\end{table}
\section{Experiment}
\subsection{Setup}
Training samples are taken from a composite comprising DAVIS-NUIUIED and OceanLab ($400$ sequences), using $128 \times 128$ crops with spatial, polarity augmentations. We optimize using Adam (initial learning rate $2\times10^{-4}$, cosine decay) for $200$ epochs~\cite{kingma2015iclr}. We also employ gradient accumulation over $\mathcal{K}=10$ forward-pass partitions of $K=10^3$ events, yielding an effective large batch size of $10^4$ events per weight update. The loss weights are $\lambda = 0.001$ and $\lambda_0 = 0.01$. 

In the absence of ground truth, we generate a reference using RAFT~\cite{teed2020raft} on consecutive frames. The input to event-based models comprises all events spanning the corresponding time windows. While this measures relative rather than absolute accuracy, it enables supervised assessments via average endpoint error (AEE~$\downarrow$) and outlier percentage ($\%_\text{3px}\downarrow$). 

For comprehensive evaluations on AquaticVision, Aqua-Eye, and the DAVIS-NUIUIED test split, we compare with FireNet and EV-FlowNet, including their SNN variants~\cite{zhu2018ev,hagenaars2021nips}. All baselines are retrained from scratch on our composite set. Finally, unsupervised metrics, variance-based flow warp loss (FWL~$\uparrow$)~\cite{stoffregen2020eccv} and time-based ratio of squared average timestamps (RSAT~$\downarrow$)~\cite{hagenaars2021nips} are reported as a supplement.

\begin{figure*}[!t]
    \centering
    \renewcommand{\arraystretch}{0.5} 
    \setlength{\tabcolsep}{1pt} 

    \subfloat[Frame]{%
        \begin{tabular}{c}
            \stackinset{r}{1pt}{t}{1pt}{%
            \includegraphics[width=0.03\textwidth]{img/ex_qual/1/scheme.png}}{\frame{\includegraphics[width=0.135\textwidth, height=50pt]{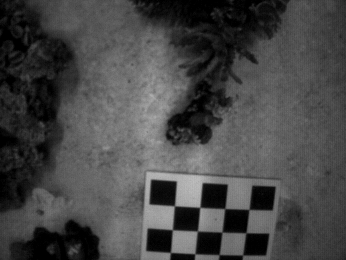}}}\\
            \frame{\includegraphics[width=0.135\textwidth, height=50pt]{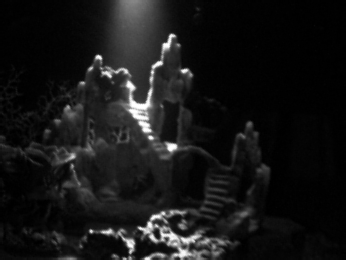}}\\
            \frame{\includegraphics[width=0.135\textwidth, height=50pt]{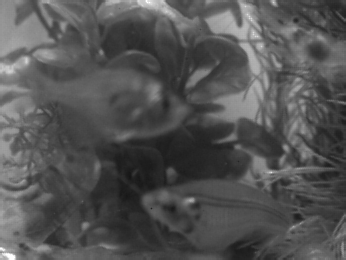}}\\
            \frame{\includegraphics[width=0.135\textwidth, height=50pt]{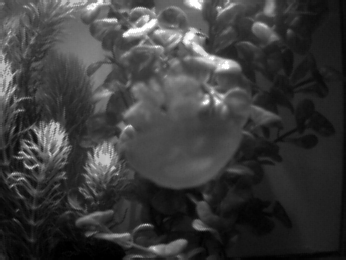}}
        \end{tabular}
    }
    \hspace{-5pt} 
    \subfloat[Reference]{%
        \begin{tabular}{c}
            \frame{\includegraphics[width=0.135\textwidth, height=50pt]{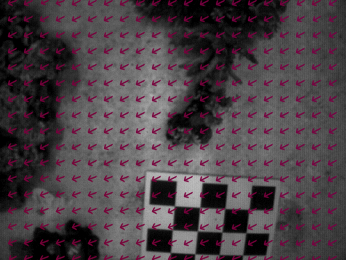}} \\
            \frame{\includegraphics[width=0.135\textwidth, height=50pt]{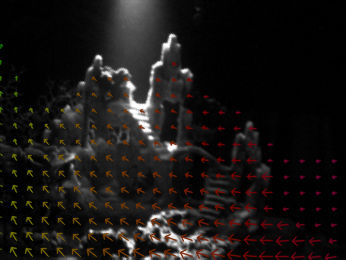}}\\
            \frame{\includegraphics[width=0.135\textwidth, height=50pt]{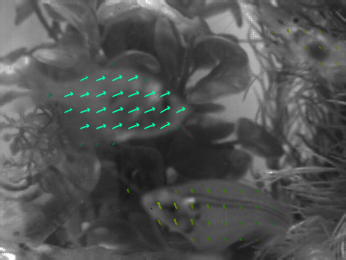}}\\
            \frame{\includegraphics[width=0.135\textwidth, height=50pt]{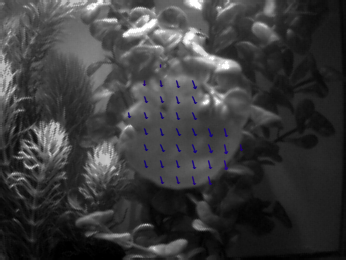}}
        \end{tabular}
    }
    \hspace{-5pt} 
    \subfloat[Events]{%
        \begin{tabular}{c}
            \frame{\includegraphics[width=0.135\textwidth, height=50pt]{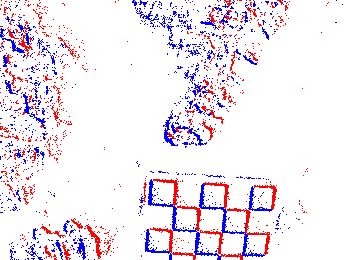}} \\
            \frame{\includegraphics[width=0.135\textwidth, height=50pt]{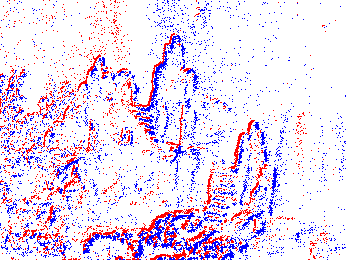}}\\
            \frame{\includegraphics[width=0.135\textwidth, height=50pt]{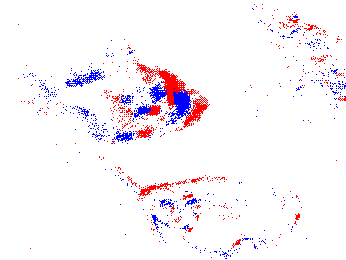}}\\
            \frame{\includegraphics[width=0.135\textwidth, height=50pt]{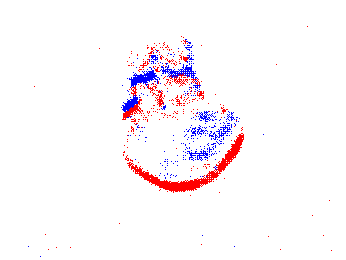}}
        \end{tabular}
    }
    \hspace{-5pt}
    \subfloat[Ours]{%
        \begin{tabular}{c}
            \frame{\includegraphics[width=0.135\textwidth, height=50pt]{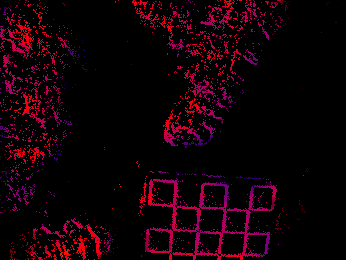}} \\
            \frame{\includegraphics[width=0.135\textwidth, height=50pt]{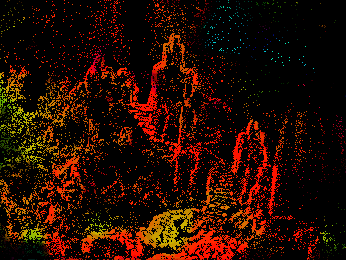}}\\
            \frame{\includegraphics[width=0.135\textwidth, height=50pt]{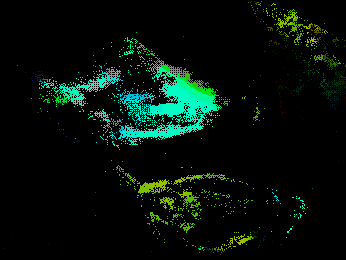}}\\
            \frame{\includegraphics[width=0.135\textwidth, height=50pt]{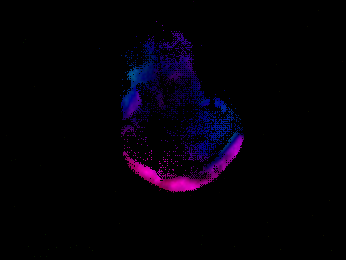}}
        \end{tabular}
    }
    \hspace{-5pt}
    \subfloat[XLIF-FireNet]{%
        \begin{tabular}{c}
            \frame{\includegraphics[width=0.135\textwidth, height=50pt]{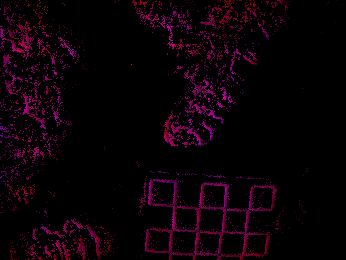}} \\
            \frame{\includegraphics[width=0.135\textwidth, height=50pt]{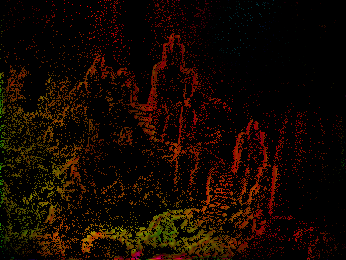}}\\
            \frame{\includegraphics[width=0.135\textwidth, height=50pt]{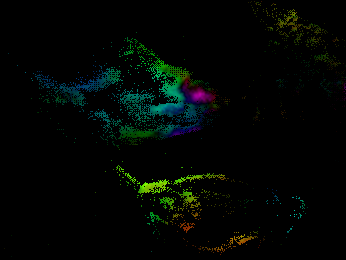}}\\
            \frame{\includegraphics[width=0.135\textwidth, height=50pt]{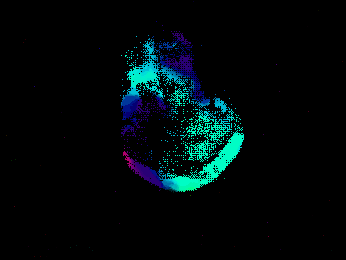}}
        \end{tabular}
    }
    \hspace{-5pt}
    \subfloat[LIF-EV-FlowNet]{%
        \begin{tabular}{c}
            \frame{\includegraphics[width=0.135\textwidth, height=50pt]{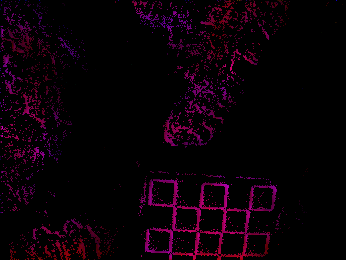}} \\
            \frame{\includegraphics[width=0.135\textwidth, height=50pt]{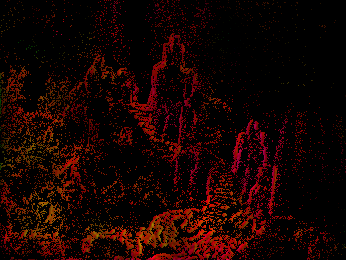}}\\
            \frame{\includegraphics[width=0.135\textwidth, height=50pt]{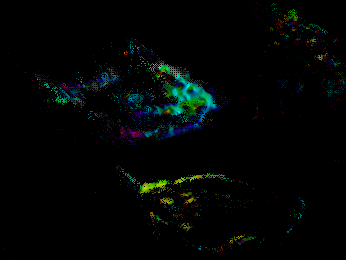}}\\
            \frame{\includegraphics[width=0.135\textwidth, height=50pt]{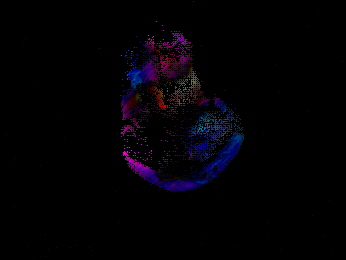}}
        \end{tabular}
    }
    \hspace{-5pt}
    \subfloat[LIF-FireNet]{%
        \begin{tabular}{c}
            \frame{\includegraphics[width=0.135\textwidth, height=50pt]{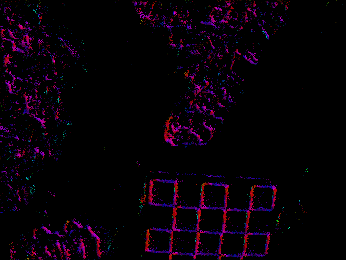}} \\
            \frame{\includegraphics[width=0.135\textwidth, height=50pt]{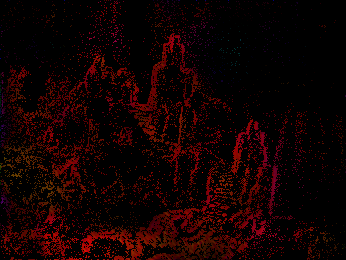}}\\
            \frame{\includegraphics[width=0.135\textwidth, height=50pt]{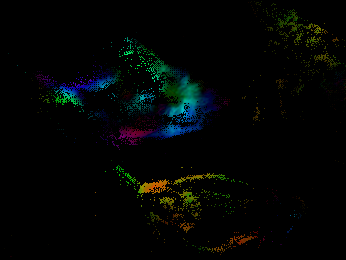}}\\
            \frame{\includegraphics[width=0.135\textwidth, height=50pt]{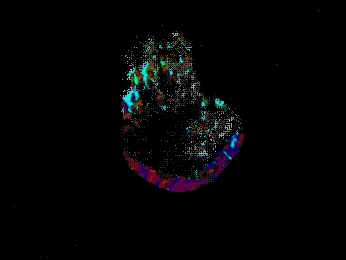}}
        \end{tabular}
    }

    \caption{Visual results on AquaticVision, DAVIS-NUIUIED, and Aqua-Eye datasets, with a color-coding scheme shown alongside.}
    \label{img/ex_qual}
\end{figure*}
\begin{table}[!t]
\centering
\footnotesize
\caption{Computational Efficiency Comparison}
\label{table/eff}
\definecolor{CEC_Best}{HTML}{FDC745}   
\definecolor{CEC_Good}{HTML}{FFDF20}   
\definecolor{CEC_Avg}{HTML}{FFF085}    
\definecolor{CEC_Poor}{HTML}{FEF9C2}   
\resizebox{\columnwidth}{!}{%
\begin{tabular}{lrrrrr}
\toprule
& {\textbf{Params.}} & {\textbf{Ops}\textsuperscript{*}} & {\textbf{Firing Rate}} & {\textbf{Power}} & {\textbf{Latency}} \\
& {(\si{M})} & {(\si{G})} & {(\si{\%})} & {(\si{mJ})} & {(\si{ms})} \\
\midrule
FireNet & 0.15 & 4.85 & ---  & 22.32 &  33.87\\
EV-FlowNet & 14.13 & 28.54 & --- & 131.28 &  61.24\\
\addlinespace
LIF-FireNet & \cellcolor{CEC_Best}0.074 & \cellcolor{CEC_Poor}2.95 & \cellcolor{CEC_Poor}30.42 & \cellcolor{CEC_Poor}2.67 &  \cellcolor{CEC_Best}23.90\\
LIF-EV-FlowNet & 20.39 & 23.73 & 41.57 & 21.33 &  47.27\\
\addlinespace
Ours & \cellcolor{CEC_Poor}0.134& \cellcolor{CEC_Best}2.31 & \cellcolor{CEC_Best}24.35 & \cellcolor{CEC_Best}2.08 &  \cellcolor{CEC_Poor}26.49\\
\bottomrule
\end{tabular}%
}
\begin{minipage}{\columnwidth}
\footnotesize
\vspace{2pt}
\textsuperscript{*}Ops (Operations) refer to FLOPs in ANNs, SOPs in SNNs.
\end{minipage}
\end{table}

\begin{figure}[t]
    \centering
    \subfloat[Isotropic scaling]{\includegraphics[width = 0.24\textwidth]{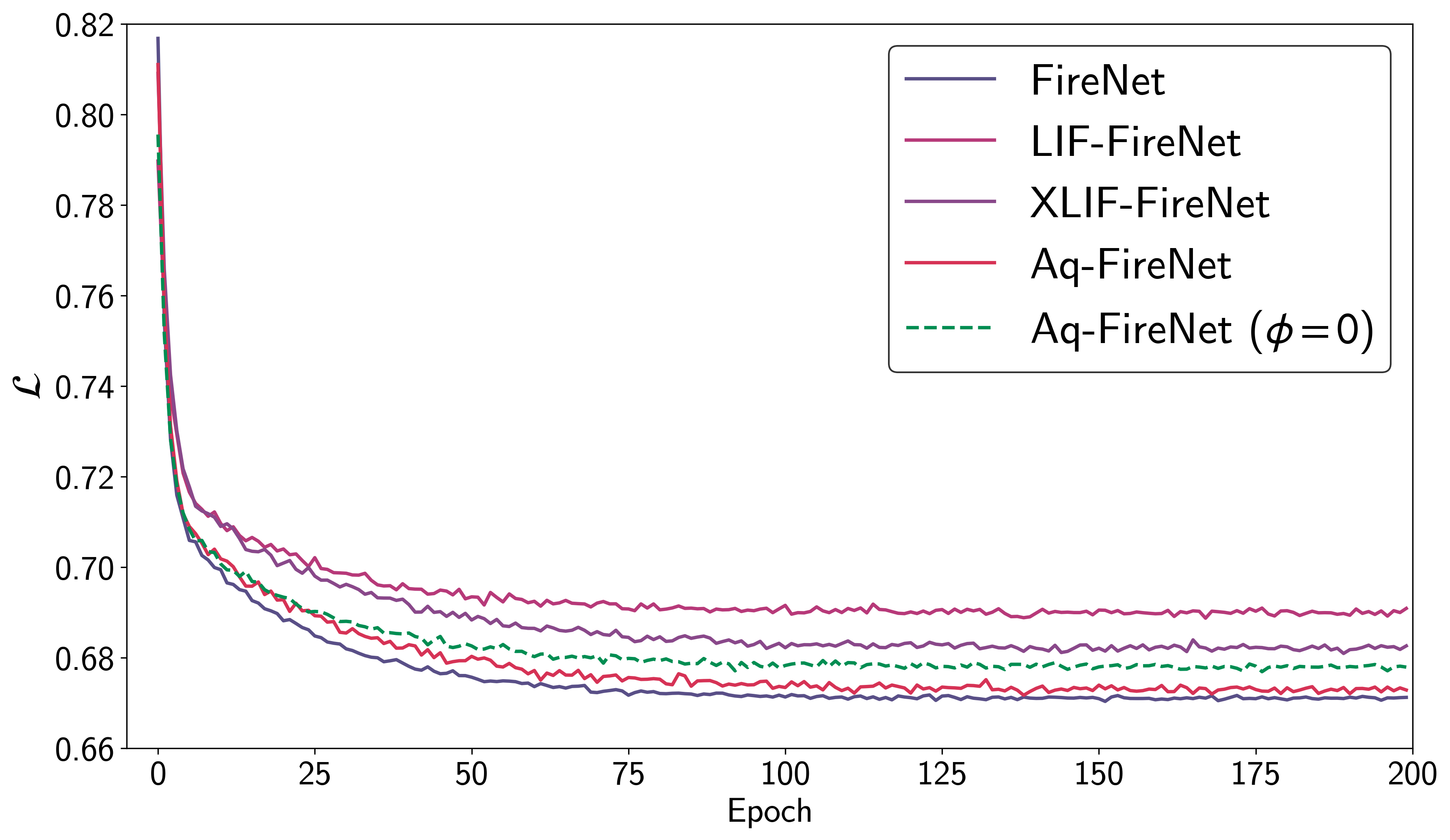}}\hfill
    \subfloat[$L_0$ regularizer]{\includegraphics[width = 0.24\textwidth]{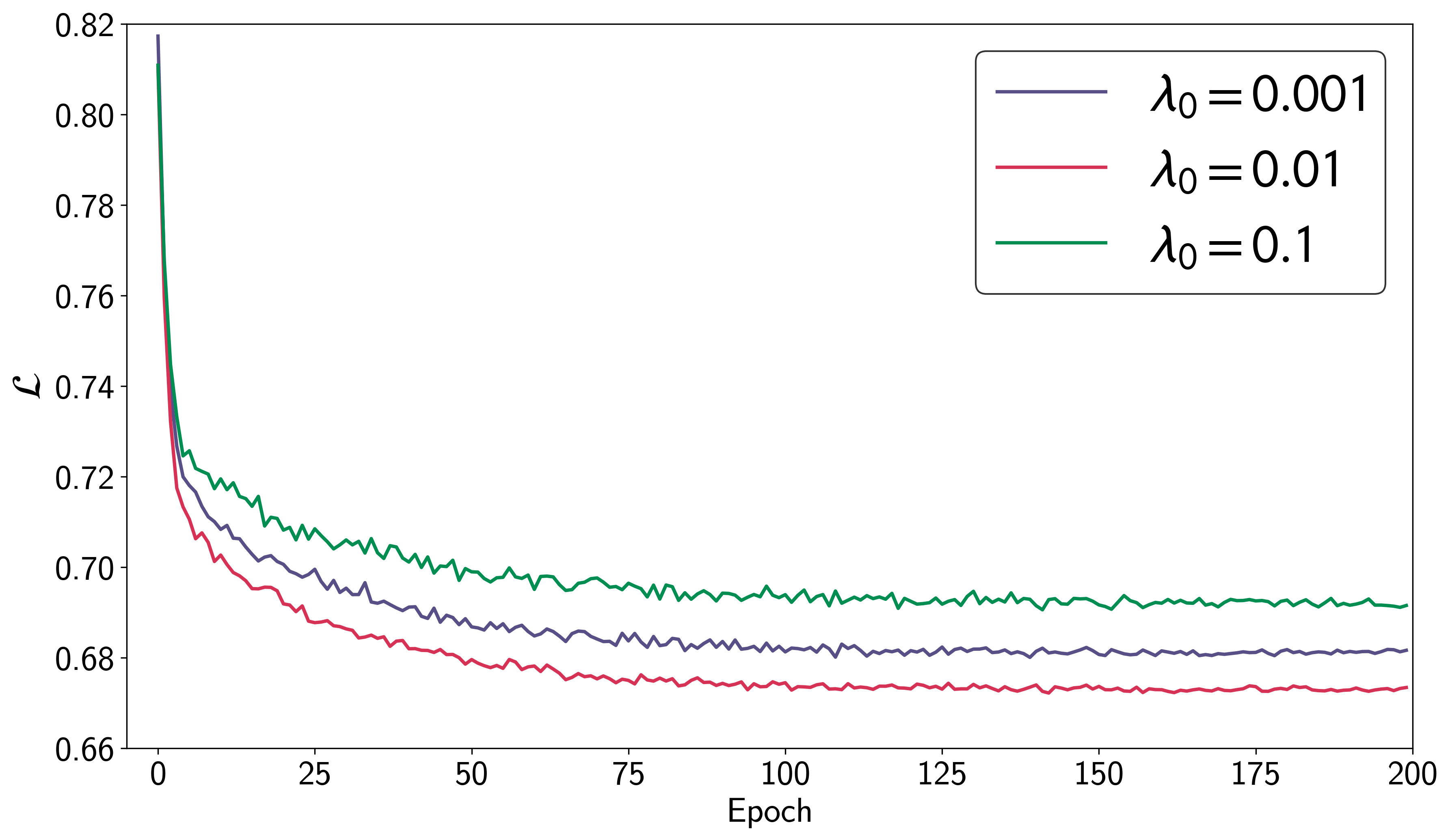}}
    \caption{Ablation study on how our modules affect learning performance.}
    \label{img/ex_learn}
\end{figure}
\begin{figure}[t]
    \centering
    \subfloat[Frame]{\frame{\includegraphics[width = 0.32\columnwidth, height=45pt]{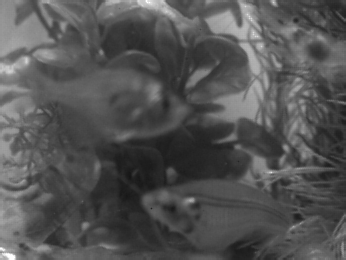}}}\hfill
    \subfloat[Without optical flow]{\frame{\includegraphics[width = 0.32\columnwidth, height=45pt]{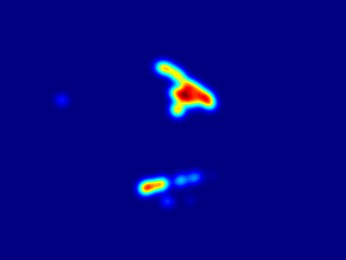}}}\hfill
    \subfloat[With optical flow]{\frame{\includegraphics[width = 0.32\columnwidth, height=45pt]{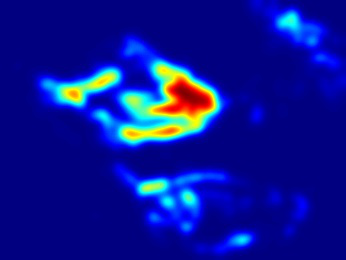}}}
    \caption{Our neuromorphic optical flow facilitates camouflaged fish detection.}
    \label{img/ex_map}
\end{figure}
\subsection{Quantitative and Qualitative Evaluation}\label{section/eval}
Tables~\ref{table/ex_quan} and~\ref{table/ex_quan2} indicate that our Aq-FireNet consistently outperforms SNN-based FireNet variants across all scenarios, and earns competitive runner-up performance against larger ANN models in most cases. Our core design is also verified by quick tests, which confirm that removing either the isotropic scaling ($\phi=0$) or the $L_0$ regularizer yields a performance drop across supervised and unsupervised metrics.

As shown in Fig.~\ref{img/ex_qual}, our method yields optical flow estimates more faithful to the reference than state-of-the-art SNN baselines across challenging aquatic scenes. It effectively handles severe visual degradations like uneven lighting and turbidity-induced low contrast. Particularly, our approach robustly tracks the slight motion of camouflaged, weakly dynamic marine life that the competitors often fail to distinguish from backgrounds.

Underwater deployment demands strict resource and latency management. Table~\ref{table/eff} reports theoretical energy consumption by distinguishing floating-point operations (FLOPs) for ANNs from synaptic operations (SOPs) for SNNs~\cite{zhou2023spikformer,kosta2023adaptive}. Our method earns the lowest computational cost, sparsest firing rate, and highest energy efficiency, while securing runner-ups in model size and inference latency. This confirms its viability for real-time, resource-constrained aquatic tasks.

As Fig.~\ref{img/ex_learn} illustrates, Aq-FireNet achieves lower loss than its linear-warping baseline ($\phi = 0$), confirming that isotropic scaling is essential for compensating complex, non-translational dynamics. Analysis of the $L_0$ regularizer reveals a trade-off in noisy aquatic contexts. Excessive regularization ($\lambda_0 = 0.1$) causes underfitting and training instability, whereas insufficient penalization ($\lambda_0 = 0.001$) fails to disentangle true motion from ambient noise and then slows convergence speed.

As visualized via feature maps in Fig.~\ref{img/ex_map}, we verify our flow estimates on motion segmentation as a sample of downstream tasks. Camouflaged marine life often remains high photometric similarity to surrounding backgrounds and nearly indistinguishable in frames. Shifting attention to changing pixels, our neuromorphic flow fields isolate dynamic contours and enable precise target extraction against cluttered aquatic backgrounds.

\section{Conclusion}
This work pioneers underwater motion field estimation via a self-supervised SNN framework. Visual and quantitative results verify its effectiveness and robustness. It opens new avenues for lightweight, real-time, and low-cost perception on resource-constrained underwater edge systems.

\bibliographystyle{IEEEtran}
\bibliography{reference}
\end{document}